\documentclass[9pt, technote]{IEEEtran}
\usepackage{cite}
\usepackage{url}
\usepackage{hyperref}
\usepackage{graphicx}
\usepackage{textcomp}
\usepackage{amsmath}
\usepackage{mathtools}
\usepackage{tabularx}
\usepackage{wasysym}

\newcolumntype{s}{>{\centering\arraybackslash\hsize=.5\hsize}X}

\DeclareMathOperator{\atan}{atan}

\begin{document}

\title{A Continuum Manipulator for Open-Source Surgical Robotics Research and Shared Development}
%
%
%

\author{Angus~B.~Clark\textsuperscript{$\dagger$},~\IEEEmembership{Graduate~Student~Member,~IEEE,}
        Visakan~Mathivannan\textsuperscript{$\dagger$},
        and~Nicolas~Rojas,~\IEEEmembership{Member,~IEEE}
\thanks{$\dagger$ Authors contributed equally to this work. Angus B. Clark, Visakan Mathivannan, and Nicolas Rojas are with the REDS Lab, Dyson School of Design Engineering, Imperial College London, 25 Exhibition Road, London SW7 2DB, UK.  (e-mail: {\tt\footnotesize
 \{a.clark17, visakan.mathivannan16, n.rojas\}@imperial.ac.uk})}
}

\maketitle

\begin{abstract}
Many have explored the application of continuum robot manipulators for minimally invasive surgery, and have successfully demonstrated the advantages their flexible design provides---with some solutions having reached commercialisation and clinical practice. However, the usual high complexity and closed-nature of such designs has traditionally restricted the shared development of continuum robots across the research area, thus impacting further progress and the solution of open challenges. In order to close this gap, this paper introduces ENDO, an open-source 3-segment continuum robot manipulator with control and actuation mechanism, whose focus is on simplicity, affordability, and accessibility. This robotic system is fabricated from low cost off-the-shelf components and rapid prototyping methods, and its information for implementation (and that of future iterations), including CAD files and source code, is available to the public on the \href{https://github.com/OpenSourceMedicalRobots}{Open Source Medical Robots initiative}'s repository on GitHub, with the control library also available directly from Arduino. Herein, we present details of the robot design and control, validate functionality by experimentally evaluating its workspace, and discuss possible paths for future development. 
\end{abstract}


\section{Introduction}
\IEEEPARstart{M}{inimally} invasive surgery (MIS) provides surgeons with a methodology that reduces patient recovery duration, pain, and infection rate due to the limited trauma and scarring caused \cite{huang2011natural}. MIS typically involves the traversal of tools along a small opening, either created using a device such as a trocar or using existing natural passages such as in natural orifice transluminal endoscopic surgery (NOTES), to the site of interest \cite{chamberlain2009comprehensive}. Typical tools include cameras, cutters, vacuums, and grippers, which are passed through a tubular controllable structure guiding the tools to the surgery site. Active research has focused on the development of continuously bending flexible structures over typical rigid tubes \cite{ouyang2016design}, as they can provide improved path following and reduced pain for the patient \cite{hu2019design}.

The development of low-cost devices for endoscopic and surgical procedures has specifically been of major interest for developing countries, where the high cost of typical devices can be restrictive \cite{campisano2017gastric, loveland2012pediatric}. The increasing access to 3D printing has introduced the design of printable devices, which can be printed at low cost and anywhere with a power source. Previous research has explored both metal \cite{hu2019design} and plastic \cite{yang20163d} printed continuum structures, usually controlled with tendon actuation. These printed backbones normally require assembly once printed, through assembly and routing tendons.

While improvements have been made to the accessibility of continuum robots, the lack of standardised approaches to continuum robot design, combined with their inherent control complexity, creates a barrier of entry to the research field \cite{walker2013continuous}. Open-source hardware developments, such as single-board microcontrollers from Arduino, have previously demonstrated the ability to overcome difficulties in adoption and their successes are attributed to being low cost, easy to use, and flexible for advanced use \cite{Zlatanov2016arduino}. The limited number of open-source surgical robots has been highlighted as a factor in the restriction of medical robot research \cite{chen2013open}.

\begin{figure}[!t]
    \centering
    \includegraphics[width=0.76\linewidth]{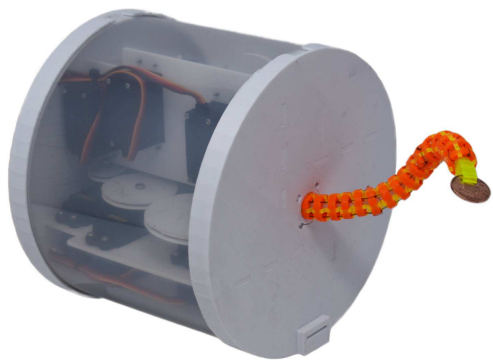}
    \caption{The resulting 3-segment ENDO continuum robot with control and actuation mechanism demonstrating the multi-curve topology achievable, and shown grasping a 1 penny coin to indicate scale.}
    \label{fig:main}
\end{figure}

Research into open-source continuum robot hardware is an unexplored field, presenting an opportunity to drive innovation \cite{burtet2018open}. Comparisons can therefore be made to other active mechatronic medical devices, where the successful implementation of open source devices has been measured and proven \cite{niezen2016open}. Therefore, due to the rapid successes of these devices, it can be assumed that using a similar structure will provide favourable results, and presents an opportunity to resolve limitations in the global adoption of continuum robots and solve open challenges to speed up commercialisation \cite{Burgner-Kahrs_review2015}. We present herein ENDO, a complete 3-DOF open-source continuum system fabricated from affordable off-the-shelf materials and equipment, including continuum robot, actuation system, and control architecture. All CAD files and source code presented are made available to the public on the Open Source Medical Robots initiative's repository on GitHub: \url{https://github.com/OpenSourceMedicalRobots}, with the control library also available directly from the Arduino Library Manager. More details can also be seen in the \href{https://youtu.be/eEjlFD5Np10}{video}.

\begin{figure}[!t]
    \centering
    \includegraphics[width=\linewidth]{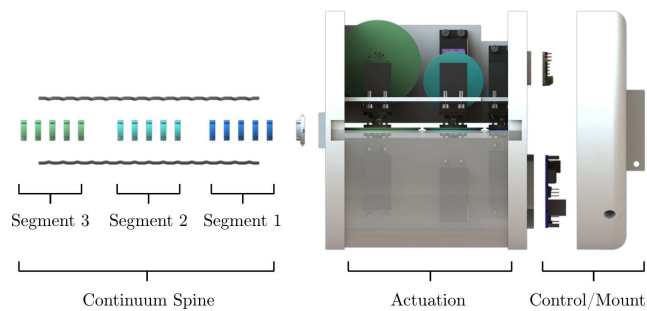}
    \caption{ENDO 3-segment continuum robot components: Spine composed of 3 segments (green, light blue, and dark blue) each with 5 rigid discs, connected with 4 TPU ligaments \textbf{(left)}; Actuation mechanism with 4 servos per segment \textbf{(centre)}; and control and mounting mechanism for attaching to a UR5 robot \textbf{(right)}.}
    \label{fig:robotrender}
\end{figure}

\section{Design and Assembly}
ENDO was designed to be manufactured using an off-the-shelf FFF 3D printer, such as the Prusa i3 MK3S, and uses affordable off-the-shelf components for the mechatronic systems. The exploded structure of the continuum robot can be seen in Fig.~\ref{fig:robotrender}, indicating the continuum spine, actuation, and control systems.

\subsection{Continuum Spine}
\begin{figure}[!t]
    \centering
    \includegraphics[width=\linewidth]{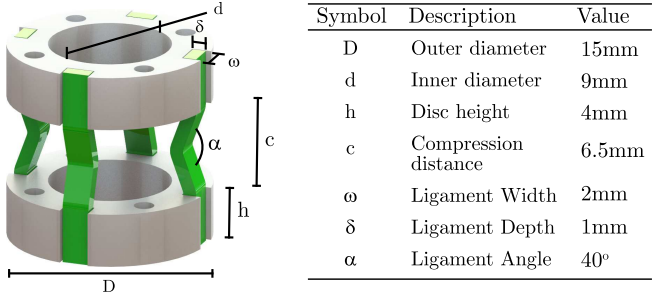}
    \caption{Subsection view between two rigid discs identifying key user changeable parameters {\bf (left)}. The description of these parameters and dimensions used in the developed robot are also shown {\bf (right)}.}
    \label{fig:spine}
\end{figure}

The spine is formed from rigid discs (15mm diameter, 4mm height) using ABS (Acrylonitrile Butadiene Styrene) and flexible ligaments (2mm $\times$ 1mm cross-section) using TPU (Thermoplastic Polyurethane), both 3D printed. The rigid discs provide guides and termination points for the tendons, and the ligaments act as a spring returning the spine to its origin when actuation is removed. A parametric design was implemented to enable adaptability depending on the user's manufacturing capabilities and desires, such as varying the spine stiffness, partially determined by the ligament angle ($\alpha$, set at 40\textdegree). These parameters defining the spine are shown in Fig.~\ref{fig:spine}.

\subsection{Actuation System}
A four tendon actuation method was selected over a trigonal tendon configuration to reduce stress on the tendons and enable lower torque motors, as well as simplify the control scheme. The mechanism was built compartmentalised and utilises the same parametric design as the spine. MG996R servo motors were selected to actuate the tendons, and a pulley system of progressively larger pulleys was developed to control each spine segment, accounting for the motion of previous segments. The radius of each pulley ($R_i$) is determined by
\begin{equation}\label{pulley}
    R_{i}\psi_{max} = \sum\limits_{j=1}^{i}S_{j, max} ,
\end{equation}
where $\psi_{max}$ is the maximum motor range, $S_{j, max}$ is the maximum segment pull length, and $i$ is the segment being controlled, with $i=1$ being the segment closest to the base.

\subsection{Control System}\label{controlsystem}
\begin{figure}[!t]
    \centering
    \includegraphics[width=\linewidth]{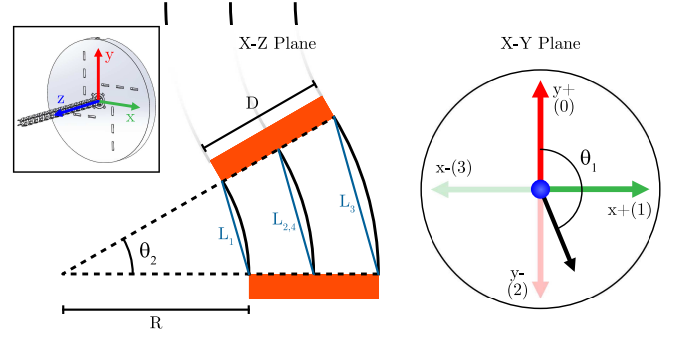}
    \caption{Parameters describing the bending of the spine: Mapping a reference Cartesian plane to the robot workspace {\bf (top left)}, bending of a sub-segment in the X-Z plane {\bf (bottom left)}, and rotation around the Z-axis, indicating the arrangement of motor channels to cardinal position (0,1,2,3) {\bf (right)}.}
    \label{fig:math}
\end{figure}

Control of the robot is provided using an Arduino Uno and a 16-channel PWM servo driver. We developed a custom Arduino library, accessible on the Arduino Library Manager as "Endo-Continuum-Robot-Library", providing user-friendly control of the robot through pre-defined functions. This enables directional control of the robot from only two angles ($\theta_1$ and $\theta_2$) for each segment, shown in Fig.~\ref{fig:math}. 

To provide a simplified level of control, we do not take into account the twist effect due to gravity. Further, we model two types of actuation: boundary condition actuation ($\frac{1}{2\pi}$, $\pi$, $\frac{3}{2\pi}$, $2\pi$) where 3 tendons are active, and actuation between these boundaries, where only 2 tendons are active. Boundary condition actuation refers to actuation in one of the 4 cardinal directions ($\pm X$ or $\pm Y$). In these scenarios the associated tendon and also the two adjacent tendons are active. Comparatively, actuation between boundaries occurs when values of $\theta_1$ are not equal to one of the aforementioned conditions. In these scenarios only 2 tendons are active.

Firstly, we model the scenario of a boundary condition actuation by assuming that the opposing tendon ($L_{3}$) remains unaltered, thus $L_{3}=L$ where $L$ is the maximum separation between two rigid discs. We compute the changes in tendon lengths $L_1$ and $L_{2,4}$ as follows:

\begin{equation}\label{1}
    \sin({0.5\theta_2})= \frac{0.5L}{D+R},
\end{equation}
where the radius of curvature $R$ is determined by:
\begin{equation}\label{3}
    R=\frac{0.5L_1}{\sin(0.5\theta_2)}
\end{equation}
Substituting equation~(\ref{3}) into equation~(\ref{1}), we have
\begin{equation}
    L_1 = L-2D\sin(0.5\theta_2).
\end{equation}
We can then solve for $L_{2,4}$, getting
\begin{equation}\label{2}
    L_{2,4} = 0.5L+0.5L_1.
\end{equation}

\begin{figure}[!t]
    \centering
    \includegraphics[width=\linewidth]{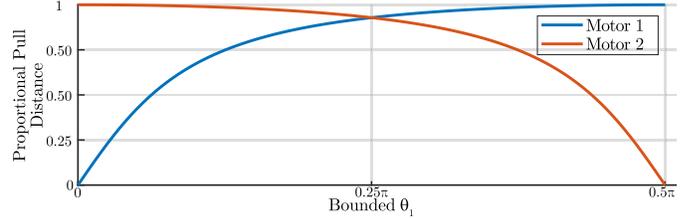}
    \caption{Graph depicting the simplified rounded square wave proportional contraction of the two tendons when actuating between boundaries, where the motors are adjacent to the desired direction $\theta_{1}$, which has been bounded to values within 0 and $\frac{1}{2\pi}$.}
    \label{fig:wave}
\end{figure}

For actuation between boundaries the active tendons are determined by the $\theta_1$ quadrant position, and we use a smoothed square wave to determine their lengths:
\begin{align}
    L_{1.1} &= \frac{L}{\atan(4)}\atan\left(4\sin(\theta_1\right),\\
    L_{1.2} &= \frac{L}{\atan(4)}\atan\left(4\cos(\theta_1\right).
\end{align}
This smoothed square wave can be seen in Fig.~\ref{fig:wave}, demonstrating the proportional pull of motor 1 ($L_1$) and motor 2 ($L_2$) over the active quadrant. Finally, to maintain tension in successive segments when a progressive segment is actuated, we compute the segment motor rotation $\psi_i$, where $S_i$ is the specific segment pull length:
\begin{equation}
    \psi_i = S_i+\sum\limits_{j=1}^{i-1} \psi_{j}.
\end{equation}

\begin{figure*}[!t]
    \centering
    \includegraphics[width=\textwidth]{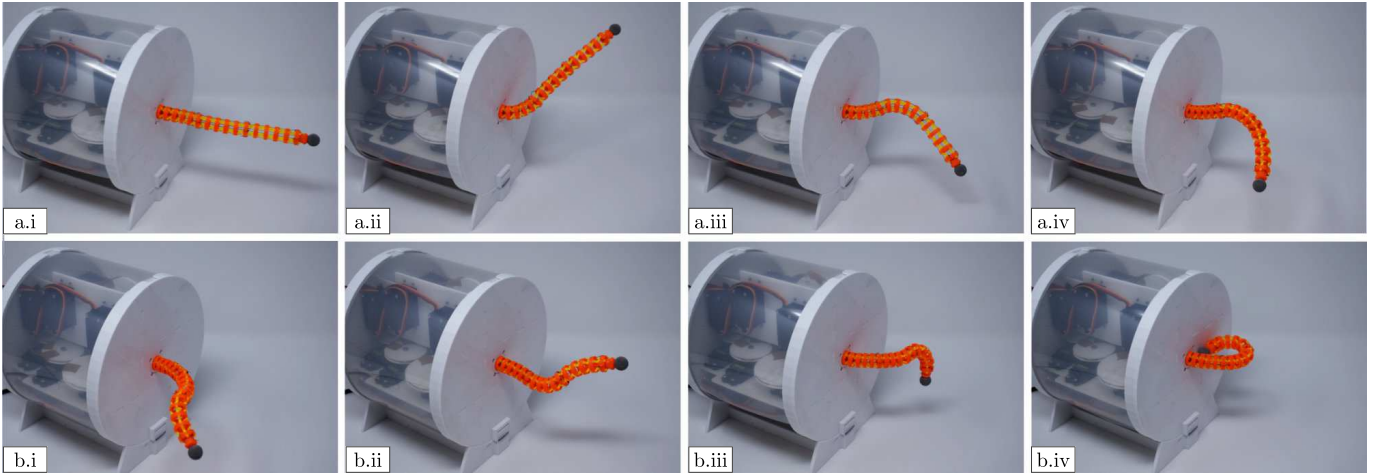}
   \caption{A series of configurations demonstrating the actuation and control system: \textbf{(a)} Successive segment actuation using no segments \textbf{(i)}, segment 1 \textbf{(ii)}, segment 1 and 2 \textbf{(iii)}, and segment 1, 2, and 3 \textbf{(iv)}; and \textbf{(b.i - iv)} example configurations utilising all 3 segments, demonstrating the reachability of the robot.}
    \label{fig:Configurations}
\end{figure*}

\subsection{End Effector}
The proposed spine design includes a hollow central channel, enabling the pass-through of tools. Typical tools include lights, grippers, and cutting tools. In addition to the continuum robot, two end effectors were also developed. The first, a simple tendon actuated monolithic gripper, shown in Fig.~\ref{fig:main} grasping a penny (£0.01, \diameter 20.3 mm), for use of the continuum robot as a hyper-redundant arm. For simplicity and strength, this was designed to be printed in a single print with no post-processing. To prevent the gripper tendon affecting the motion of the spine, a PTFE (polytetrafluoroethylene) tube providing a low resistance but fixed length traversal guide down the centre of the spine was used. In addition to this, a motion tracking mount was also created, enabling the tracking of the robot end-effector position.

\section{Evaluation}
To demonstrate the viability of this continuum system, positional experiments were performed to evaluate aspects such as the workspace size and its redundancy. The layered design can be seen functioning in Fig.~\ref{fig:Configurations}a, where the actuation of successive segments is performed while the actuation of the former segments are maintained. Further 3-segment configurations, where all 3 segments are actuated, are shown in Fig.~\ref{fig:Configurations}b demonstrating the varied reachability of the robot. Specifically, Fig.~\ref{fig:Configurations}b.iv shows the ability to explore within the workspace, rather than only on a spherical surface. All of these positions are achieved using the developed Arduino library, taking $\theta_1$ and $\theta_2$ as inputs for each segment.

To quantitatively evaluate the performance of the robot, the workspace was explored and measured using 4 motion tracking cameras (OptiTrack Flex 3, calibration error 0.166mm) and a marker fixed to the end effector of the robot. The robot was positioned by actuating through all possible segment values of $\theta_1$ and $\theta_2$ successively, with restrictions applied to prevent the spine or end effector from colliding with the base (-Z positions). 1728 motions were explored, with a run-time of approximately 1 hour. The results of the workspace exploration can be seen in Fig.~\ref{fig:workspace}, overlaid with a CAD model of the spine for size reference.

\begin{table}[!t]
    \caption{Bill of Materials}
    \label{costs}
    \begin{center}
        \begin{tabularx}{\columnwidth}{Xsss}
            \hline
            Part            & Quantity  & Cost \\
            \hline
            PETG 3D Printer Material    & 792g  & £19.80 \\
            TPU 3D Printer Material     & 2g    & £0.05  \\
            MG996R Servo Motor          & 12    & £53.10 \\
            Arduino Uno                 & 1     & £17.80 \\
            16ch PWM Servo Driver       & 1     & £9.99  \\
            \hline
            Total Cost: & - & £100.74 \\
            \hline
        \end{tabularx}
    \end{center}
\end{table}

\section{Discussion}
\begin{figure*}[!t]
    \centering
    \includegraphics[width=0.86\textwidth]{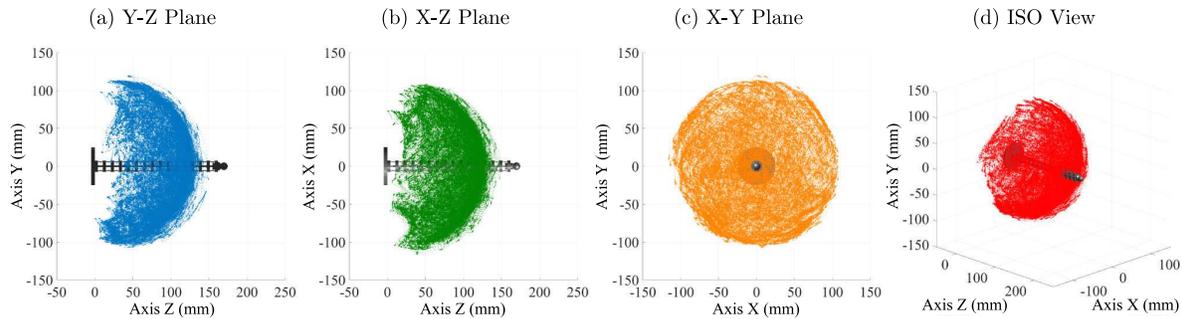}
    \caption{Experimental results showing the workspace of the continuum robot: \textbf{(a)} Y-Z plane, \textbf{(b)} X-Z Plane, \textbf{(c)} X-Y Plane, and \textbf{(d)} ISO view of the workspace.}
    \label{fig:workspace}
\end{figure*}

From the workspace results we see a strong distribution of points generating a hemispherical workspace. There is an observable uniformity in the results, which is clearly seen in Fig.~\ref{fig:workspace}c. Fig.~\ref{fig:workspace}a demonstrates that a slight skew towards the -Y axis, however it is unclear whether the cause of this distortion is due to gravity or inconsistent zeroing of the tendons. 

As the proposed design utilises tendons, the robot assembly involves a manual zeroing process to configure the tendons such that each segment is in tension at motor position 0. The manual aspect of this task presents a number of potential inaccuracies due to human errors because it is difficult to ensure each segment is equally in tension. The selection of servo motors for the robot provides an additional limitation due to the limited rotation capability when paired with the tendon zeroing process, as the pulley and motor require optimal alignment to enable to maximum rotation of the motor. While this was partially accounted for by using larger motor ranges than required, this could result in limitations at high $\theta$ values and needs to be considered for continuum robots with greater numbers of segments. This can be observed in Fig.~\ref{fig:Configurations}a and Fig.~\ref{fig:Configurations}b, where gaps in the workspace can be seen close to the faceplate and the origin.

Finally, it became apparent through repeated actuation that fatigue of the TPU ligaments occurred, affecting the manipulators workspace as the ligaments were unable to return to their full original length. This can be seen in Fig.~\ref{fig:workspace} where the initial spine length is overlaid on the workspace. While this can be accounted for by utilising thicker ligaments, this results in a higher stiffness and therefore a higher motor torque requirement, therefore a balance between the two is required. Alternative spring designs or materials could be explored to avoid this in the future.

Beyond the quantitative results of the robot, the validity of the system as an affordable open-source project can be assessed. The entire continuum robot system can be assembled from off-the-shelf components and 3D printed parts, with a total cost of approximately £100. The bill of materials for the robot with the specific component costs can be seen in Table~\ref{costs}. The printing and processing duration of 3D printed parts is not considered, as it is dependent on the printer type and print settings. Assembly of the robot, including wiring, can be completed in roughly 4 hours by a proficient user. The user friendly nature of the Arduino platform, combined with the developed open-source library, enables the assembled continuum manipulator to be quickly operational following assembly.

While the library provides users with an easy method for control, this is limited by the assumptions taken to produce a simplified control system. For achieving more complex end-effector positions, a more accurate control system would be needed. One of the goals of this project was to produce a platform specifically for the development of continuum robots, where evaluating control schemes would be an expected future direction. The provided simple control scheme allows a user to validate their assembled robot, and acts as a starting base for improvements. Furthermore, the included parametric design allows the user to alter key design variables, generating a robot to suit their needs. This reduces the skill requirement of CAD for future users and creates a larger potential audience.

\section{Future Developments}
The open source nature of the design provides a great opportunity to utilise expertise across multiple fields, including medical, mechanical, electronic, and control disciplines, as it enables easier collaboration towards solving complex problems. Future developments can be conducted by multiple collaborators through the GitHub's pull request feature. These future developments could involve resolving  challenges to enable future commercialisation to clinical practice, namely instrumentation, visualization, and integration; Human–Machine Interaction; and shape and force sensing \cite{Burgner-Kahrs_review2015}.

One key area of the current robot system which can be improved is the actuation mechanism. Presently, the design makes use of 4 sets of pulleys arranged adjacent to each other. This design presents issues in the enlargement of segments or the addition of more segments due to the relationship of the pulleys to the successive segments, resulting in the progressive growth of the pulley radii. One approach for solving this issue is regarding the motor selection, as with the use of continuously rotating motors the pulley diameter restriction is lifted, however this introduces questions regarding the control of the motor positioning and tendon zeroing.



\section{Conclusion}
In this paper we presented the development of an affordable, open-source, continuum robot system, called ENDO, to provide a platform for continuum robot research, including continuum spine, actuation system, and control system, removing the barrier of entry to the field. The proposed design is low cost with a total part cost of £100, using easily acquirable components and materials, and is relatively easy to assemble. Alongside the mechatronic systems of the robot, we also introduced an Arduino Library for the control of the robot, providing simplified control of the manipulator and a starting point for future development. We evaluated the developed robot through a workspace exploration using motion tracking, with results demonstrating a uniform distribution of points in a hemisphere workspace. Both the developed ENDO continuum robot system and Arduino control library are available on the Open Source Medical Robots GitHub (\href{https://github.com/OpenSourceMedicalRobots}{https://github.com/OpenSourceMedicalRobots}). Ultimately, this paper presents a baseline continuum robot to enable future researchers and developers to further enhance the technology. Future developments were discussed and may include improved control algorithms, redesigning aspects of the actuation mechanism, or incorporating external factors such as gravity in to the control.

\bibliographystyle{IEEEtran}
\bibliography{main}

\end{document}